# Environment for the Design and Automation of New CDPR Architectures

Josue Rivera, Julio Garrido*, Enrique Riveiro, Diego Silva

Vigo University, Vigo, Spain

```
*jgarri@uvigo.es
```

**Abstract.** This paper presents a design and automation environment to study the control trajectory for new CDPR architectures, for instance CDPRs with an unusual number of cables or different motor location in the robot frame. In order to test the environment capabilities, an architecture of a planar under-constrained CDPR was designed, simulated, and implemented using standard industrial hardware. Both the simulated model and industrial prototype were running the same trajectories to determine the time delay and the error position between them. The tests have demonstrated that the simulated model of the CDPR reproduces the trajectories of the equivalent industrial prototype with a maximum deviation of 0.35% under loading and different speed conditions, despite the time delays produced by the data transmission and the non-deterministic communication protocols used to connect the industrial automation controller with the simulated model. The results have shown that the environment is suitable for trajectory control and workspace analysis of new CDPR architectures under different dynamic conditions.

## 1 Introduction

Cable-driven parallel robots (CDPR) are characterized by employing cables to control the position and orientation of the end-effector. Since their proposal by Landsberger and Sheridan in the mid-1980s [1], the evolution of CDPRs has been continuous, with early developments such as the SkyCam, RoboCrane and FAST [2] to more recent ones such as IPAnema [3] and FASKIT [4]. Unlike conventional parallel manipulators, which can exert tensile and compressive forces on loads, CDPRs can only exert tensile forces due to the use of cables. However, cables provide CDPR with lower inertia, higher payload-to-weight ratio, higher dynamic performance, larger workspace, and higher modularity and reconfigurability [5, 6].

The literature classifies CDPRs according to several criteria, including the working space and the ratio between the number of cables $m$ and the degrees of freedom (DoF) $n$. According to the workspace, CDPRs are classified as either planar and spatial due to their motion in the plane or in space. According to the relationship between $m$ and $n$, CDPRs are classified as fully constrained and under-constrained. A CDPR is considered fully constrained when it is impossible to change its position and orientation without changing the length of its cables. In general, this architecture requires at least one more cable than degrees of freedom ($m \geq n + 1$) [7]. Any other architecture that is not fully constrained is considered under-constrained.

The workspace of a CDPR is the set of points the end effector can reach. In addition to the dimensions of the robot frame, the workspace of a CDPR is also limited by other conditions [8]: Controllable end-effector, positive tension between a minimum and maximum value to prevent the cables from breaking or sagging, and cable collision avoidance.

There is broad research into CDPR architectures to accommodate robots for different use scenarios, for instance: measurement [9], rescue [10], and maintenance [11]. Tools to test the kinematics and dynamics of new robots without the need to build a physical model to validate the concept are useful for their design phase. In a similar way, this article presents an environment to design and automate new CDPR architectures (for instance CDPRs with an unusual number of cables or different motor location in the robot frame). To validate the environment's capability to replicate the motion trajectories of a CDPR taking into account the effect of gravity, tool weight, and motion uncertainties caused by an under-constrained configuration, a 2-DoF CDPR is designed and simulated within the environment to compare its behavior with a prototype implemented with industry-standard hardware.

## 2    Kinematic modelling and workspace study of CDPR

Kinematics deals with the study of the geometry of motion, without considering the forces involved. In order to obtain the 2-Dof CDPR kinematic model, the cables were considered to be mass-less and non-elastic, as shown in Fig. 2. In the model, $O$ represents the absolute frame of the system, $p$ is the vector position of the mobile frame $P$ located in the center of the end effector, with respect to the absolute frame, $b_i$ is the vector position of the attachment point $B_i$ with respect to the mobile frame, $a_i$ is the vector position of the coiling system output point $A_i$ of the cable $i$, and $l_i$ is the length of the cable. Thus, the general expression that describes the vector close-loop for the CDPR is:

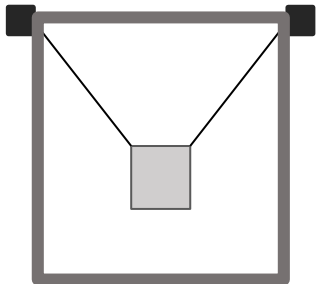 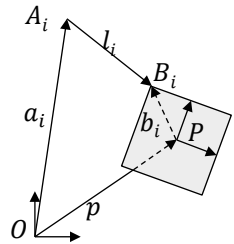

**Fig. 1.** 2-DoF CDPR model      **Fig. 2.** Geometry and kinematics of a planar cable robot.

$$a_i + l_i = p + Rb_i \tag{1}$$

Where R is the rotation matrix with respect to the absolute frame.

The backward kinematic computes the individual cable lengths from the end effector pose. A planar under-constrained CDPR can only control its position but not its orientation. Therefore, R becomes the identity matrix. Thus, the lengths of the cables are:

$$l_1 = \sqrt{\left(p_x - \frac{B_x}{2} - A_{1x}\right)^2 + \left(p_y + \frac{B_y}{2} - A_{1y}\right)^2} \tag{2}$$

$$l_2 = \sqrt{\left(p_x + \frac{B_x}{2} - A_{2x}\right)^2 + \left(p_y + \frac{B_y}{2} - A_{2y}\right)^2} \tag{3}$$

Where $B_x$ and $B_y$ are the width and height of the end effector. Also $A_{1x}$, $A_{1y}$, $A_{2x}$ and $A_{2y}$ are the position coordinates of the coiling system output. Note that $A_{1y}$ and $A_{2y}$ are the same.

The forward kinematic consists of determining the end effector pose as a function of the lengths of the cables $l_i$. The solution of the forward kinematic is a system with as many equations as the number of cables and unknowns equal to degrees-of-freedom of the system. In a 2-Dof CDPR, the forward kinematic solution can be obtained by solving Eq. 2 and Eq. 3 for $p_x$ and $p_y$ as follow:

$$p_x = \frac{A_{1x}^2 - A_{2x}^2 + B_x A_{1x} + B_x A_{2x} - l_1^2 + l_2^2}{2(A_{1x} - A_{2x} + B_x)} \tag{4}$$

$$p_y = A_{1y} - \sqrt{l_1^2 - \left(p_x - \frac{B_x}{2} - A_{1x}\right)^2} - \frac{B_y}{2} \tag{5}$$

The static equilibrium workspace analysis returns the set of positions and orientations that the end effector can reach statically (only considering the gravity effects) [12]. According to the Newton-Euler law, the equation that relates the wrenches on the end-effector (forces and moments) and the cable tensions can be written as:

$$A \cdot T = B \tag{6}$$

Where $A$ is the transpose of the Jacobian Matrix $-J^T$. $T$ is the cable tension matrix, and $B$ is the matrix of wrenches generated due to the cable actuation.

$$B = \begin{bmatrix} f_x & f_y & f_z & M_x & M_y & M_z \end{bmatrix}^T \tag{7}$$

$$T = [T_1 \ T_2 \ ... \ T_n]^T \tag{8}$$

$$A = -J^T = \begin{bmatrix} u_1 & u_2 & ... & u_n \\ c_1 \times u_1 & c_2 \times u_2 & ... & c_n \times u_n \end{bmatrix} \tag{9}$$

Where $u_i$ is the unit vector along the cable direction, and $c_i$ is the vector position of the cable attachment point to the end effector center.

From Eq. 8 we can compute the tension in each cable which is compared to the lower and upper tension limit of the system, dictated by the maximum tension supported by the cable or the maximum torque available in the motors.

The base frame of the 2-DoF CDPR used is a square of 1500mm each side, the end effector is a square of 120mm each side, and weight of 1kg. The tension limits are $0 \leq T \leq 20\ N$ due to the maximum torque of the motors. The tension map in the cables and the static equilibrium workspace of the CDPR is shown in Figs. 3 a) and b), respectively.

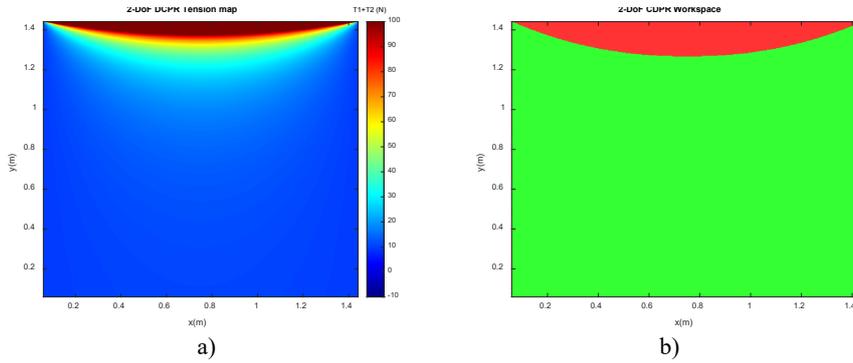

**Fig. 3.** a) Tension map in the cables. b) Static equilibrium workspace of 2-DoF CDPR

## 3   Design and automation environment for trajectory control of new CDPR architectures

Fig. 4 shows the diagram of the design and automation environment. It allows the user to test the trajectory by establishing a control loop between the controller and the simulated model (loop A) or the industrial prototype (loop B). Fig. 4 also shows the design and validation processes inside the dotted squares. The design of the simulated model has the workspace, motion, and load limits as inputs, which later translate into the robot's structure, motors and sensors, and end-effector specification.

In the framework of this paper, the industrial automation software TwinCAT 3 from Beckhoff (TwinCAT onwards) is used to control the new CDPR architectures. This software is employed to design the control trajectory strategy for testing the motion of both the simulation model and industrial prototype employing the kinematic modelling developed in section 2.

The simulation model was developed in CoppeliaSim, a robotic simulation environment that allows the user to control the behavior of each model object through scripts from inside the simulator or externally from an application programming interface (API) [13]. In the environment, the API is used to connect the simulated model to the industrial controller using Python as a gateway. On the one hand, Python and CoppeliaSim communicate using the Websocket protocol on top of the TCP protocol. On the other hand, TwinCAT and Python communicate using the ADS protocol (Automation Devise Specification) [14] on top of the TCP/IP protocol.

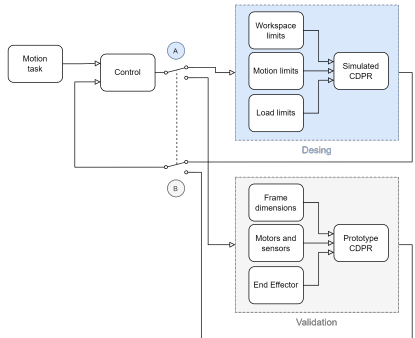
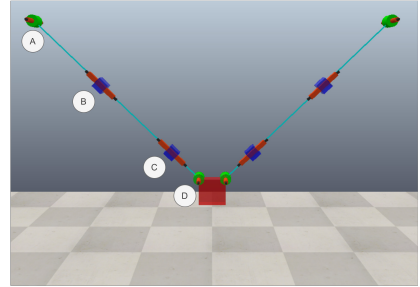

**Fig. 4.** Diagram of the design environment for trajectory control of new CDPR architectures.

**Fig. 5.** Simulation model of a CDPR in CoppeliaSim.

For the 2-DoF CDPR, a sequence of rigid bodies and joints was used without considering the coiling system and the cable sag as shown in Fig.5. The pulleys were modeled using rotational passive joints and rigid bodies (A in Fig. 5). The cables were modeled as a sequence of a prismatic joint, a rigid body, a prismatic joint, and a rigid body. The first prismatic joint is used to change the cable length (B in Fig. 5), and the second prismatic joint is used to avoid compressive stresses over the end effector (C in Fig. 5), acting as a free joint in one direction (compressive) and locking in the other direction (traction) as in [15]. To attach the end of the cable to the end-effector a similar construction to the pulley is used (D in Fig. 5).

Finally, the prototype is implemented using a Beckhoff CX5130 industrial PC with distributed periphery connected by an EtherCAT Fieldbus to compact servo motor terminals (EL7211) and compact servo motors with holding brake (AM8112).

## 4    Results

Two tests were performed to check the behavior of the designed model considering the end effector in loaded condition and two different velocities, 100mm/s and 1000mm/s. The first test compared the simulated model and industrial prototype axis positions against the target axis positions to calculate the time delay in the execution of trajectories, as visualized in Fig. 6. This figure represents the axis 1 trajectory plot of the model and prototype at high speed. Table 1 shows the time delay results from this test.

Fig. 6 indicates that the simulated axes reproduce the target trajectory and its motion corresponds to that of the real prototype axes, despite the relative time delay between the systems. The origin of the simulated time delay was the data transfer time between TwinCAT and CoppeliaSim through python and the variability introduced to the environment by the non-deterministic communication protocols available.

The second test evaluated the simulated model and industrial prototype end-effector position against the target position while executing a square trajectory to observe the difference in the model and prototype dynamics due to the end-effector weight. The position error results with respect to the target position at high speed is given in Fig. 7. Table 2 shows the average target positions and the average position of the simulated model and industrial prototype in both upper and lower horizontal trajectories at different velocities. These results show that both systems have a position error difference of less than 0.06% at low speed and 0.35% at high speed. These position errors indicate the capabilities of the environment to reproduce the dynamics of the 2-DoF CDPR with a tolerance of less than 5.52mm in the robot workspace presented in chapter 2. Therefore, the results show that the environment is suitable for developing trajectory control strategies and workspace analysis considering different dynamic conditions.

**Table 1.** Time delay of the simulated model and industrial prototype axis trajectories.

| Velocity [mm/s] | Simulated axis time delay | | Industrial prototype time delay | |
|---|---|---|---|---|
| | Axis 1 [ms] | Axis 2 [ms] | Axis 1 [ms] | Axis 2 [ms] |
| 100 | 150 | 120 | 20 | 30 |
| 1000 | 130 | 110 | 20 | 20 |

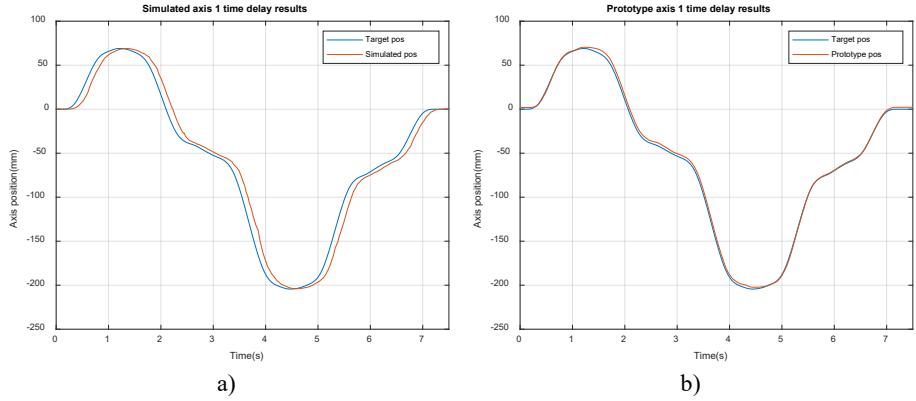

a)      b)

**Fig. 6.** Time delay between the position of the simulated and prototype axes with respect to the target position at high speed. a) Simulated axis 1. b) Prototype axis 1.

**Table 2.** Error position in simulated model and industrial prototype horizontal trajectory.

| Velocity [mm/s] | Simulated model | | | Industrial prototype | | | Error Difference [%] |
|---|---|---|---|---|---|---|---|
| | Mean target position [mm] | Mean simulated position [mm] | Position error [%] | Mean Target position [mm] | Mean prototype position [mm] | Position Error [%] | |
| 100 | 947.613 | 946.016 | 0.168 | 947.613 | 948.670 | 0.112 | 0.057 |
| | 749.578 | 750.183 | 0.081 | 749.582 | 748.844 | 0.098 | -0.017 |
| 1000 | 948.046 | 945.437 | 0.275 | 948.035 | 945.879 | 0.227 | 0.048 |
| | 749.594 | 750.510 | 0.122 | 749.600 | 746.090 | 0.468 | -0.346 |

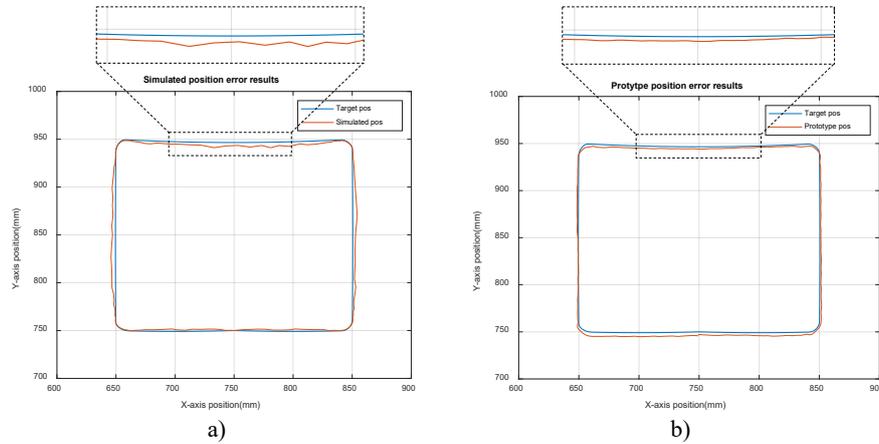

**Fig. 7.** Position error of the simulated and prototype axes with respect to the target position at high speed. a) Simulated axes. b) Prototype axes.

## 5     Conclusions

This paper examines a design and automation environment for the trajectory control of new CDPR architectures. To test the environment, a 2-DoF CDPR was designed, simulated, and implemented using industry-standard hardware to evaluate the designed CDPR and the trajectory control performance by comparing both the simulated model and industrial prototype.

The tests have shown that the simulated CDPR reproduces the end-effector movements of the prototype with a maximum deviation of 0.35% under loading conditions and at different speeds despite the time delay between both systems. These features make the environment an appropriate tool to perform motion and workspace studies of new CDPR architectures under different dynamic conditions. In addition, it allows such architectures to be validated against an industrial hardware robot by translating the design inputs (workspace, movements, and load limits) into the robot's structure, motor and sensors, and end-effector manufacturing specifications.

The current version of the environment simulates CDPRs with mass-less and non-elastic cables. Future research aims to improve the cable modeling to consider those simplifications but also cable sagging to study CDPR with a bigger workspace. Additionally, improvements in the communication will be sought to reduce the time delay between the prototype and the simulated model.